\documentclass[conference]{IEEEtran}
\IEEEoverridecommandlockouts
\usepackage{cite}
\usepackage{amsmath,amssymb,amsfonts}
\usepackage{algorithmic}
\usepackage{latexsym}
\usepackage{textcomp}
\usepackage{booktabs}
\usepackage{graphicx} 
\usepackage{xcolor}
\def\BibTeX{{\rm B\kern-.05em{\sc i\kern-.025em b}\kern-.08em
    T\kern-.1667em\lower.7ex\hbox{E}\kern-.125emX}}
\usepackage{amsmath}
\usepackage{graphicx}
\usepackage{booktabs} 
\usepackage{hyperref} 
\usepackage{url} 

\usepackage{fancyhdr}
\begin{document}

\title{LVLM-Composer's Explicit Planning for Image Generation}

\author{Spencer Ramsey, Jeffrey Lee, Amina Grant\\
Northern Caribbean University
}

\maketitle
\thispagestyle{fancy} 

\begin{abstract}
The burgeoning field of generative artificial intelligence has fundamentally reshaped our approach to content creation, with Large Vision-Language Models (LVLMs) standing at its forefront. While current LVLMs have demonstrated impressive capabilities in text-to-image generation, they often falter when confronted with complex textual descriptions demanding precise compositional understanding and visual planning. This limitation particularly impacts the accurate rendering of multiple objects, their attributes, spatial relationships, and specific poses within intricate scenes, as evidenced by benchmarks like LongBench-T2I. To address these challenges, we introduce \textbf{LVLM-Composer}, a novel 10-billion parameter scale LVLM specifically engineered for enhanced compositional image synthesis. Our method incorporates a \textbf{Hierarchical Semantic Planning Module} for structured prompt decomposition and a \textbf{Fine-Grained Feature Alignment Mechanism} for precise visual guidance during generation. We propose a multi-stage training paradigm, featuring Hierarchical Semantic-Visual Grounding Pre-training and Compositional Planning Reinforcement Learning with Self-Correction, to instill robust compositional reasoning. Extensive experiments on the LongBench-T2I benchmark, utilizing automatic evaluation by Gemini-2.0-Flash and InternVL3-78B, demonstrate LVLM-Composer's superior performance across critical compositional dimensions including object accuracy, composition fidelity, and pose accuracy, significantly outperforming state-of-the-art baselines. An in-depth ablation study further validates the indispensable contribution of our proposed modules, while human evaluations confirm the perceptual superiority of our generated images. LVLM-Composer represents a significant step towards truly controllable and compositionally accurate open-ended text-to-image generation.
\end{abstract}

\section{Introduction}
The rapid advancements in artificial intelligence have brought forth a paradigm shift in various domains, none more captivating than the realm of generative AI. Within this landscape, \textbf{Large Vision-Language Models (LVLMs)} have emerged as powerful agents, demonstrating impressive capabilities in understanding and generating content across modalities. Specifically, the application of LVLMs to generative tasks, particularly \textbf{image generation} and increasingly video generation \cite{zhou2024less}, has revolutionized the way we interact with creative processes, enabling the synthesis of photorealistic and semantically rich images from mere textual descriptions. This remarkable progress holds immense significance, promising transformative impacts on digital content creation, design, virtual reality, and even scientific visualization, by democratizing access to complex visual synthesis.
Despite these significant strides, current LVLMs for image generation, particularly lightweight models like Janus-Pro-1B \cite{JanusPro1B}, still face substantial \textbf{challenges} when confronted with sophisticated textual prompts demanding high levels of \textbf{compositional understanding} and \textbf{visual planning}. Generating images that accurately depict multiple objects with precise attributes, intricate spatial relationships, specific poses, and consistent stylistic elements within a complex scene remains a formidable hurdle. Existing models often struggle to maintain coherence across various visual components, leading to inconsistencies in object placement, attribute misalignments, or a failure to capture the nuanced interplay between different elements described in the prompt. This limitation largely stems from the inherent difficulty in explicitly teaching models to perform structured visual planning and fine-grained alignment between textual semantics and visual features during the generation process. The need for better compositional abilities is underscored by the development of holistic benchmarks designed to test complex, instruction-based generation and editing \cite{zhou2025draw, wang2025complexbench}. The "LongBench-T2I" benchmark \cite{zhou2025draw}, utilizing advanced evaluators like Gemini-2.0-Flash and InternVL3-78B, vividly illustrates these existing gaps, pushing the boundaries of what is expected from compositional text-to-image models.
Our primary \textbf{motivation} is to directly address these compositional generation limitations by developing an LVLM that can effectively "think" in terms of visual plans and execute them with high fidelity, aiming to push towards stronger generalization across multiple capabilities \cite{zhou2025weak}. We aim to move beyond merely generating plausible images to consistently producing images that accurately reflect complex, multi-faceted textual descriptions. By doing so, we seek to empower users with unprecedented control over the generated visual content, unlocking new possibilities for creative expression and practical applications. We hypothesize that by explicitly integrating mechanisms for hierarchical semantic parsing—a form of divide-and-conquer strategy for complex tasks \cite{zhu2025divide}—and fine-grained visual alignment within the LVLM's architecture and training regimen, we can significantly enhance its ability to deconstruct complex textual prompts, unraveling their chaotic context \cite{zhou2023thread}, into actionable visual instructions and then synthesize images that precisely adhere to these plans.
In this paper, we propose \textbf{LVLM-Composer}, a novel 10-billion parameter scale LVLM specifically designed to excel in compositional text-to-image generation. Our approach introduces a \textbf{multi-stage training paradigm} that systematically imbues the model with advanced visual planning capabilities. The first stage focuses on \textbf{Hierarchical Semantic-Visual Grounding Pre-training}, where the LVLM learns to explicitly map complex textual descriptions to fine-grained visual elements, their attributes, and their spatial relationships using richly annotated datasets. This involves the novel use of "planning tokens" as an intermediate representation, compelling the model to articulate its visual strategy. The second stage employs \textbf{Compositional Planning Reinforcement Learning with Self-Correction} \cite{Nachum2019HierarchicalRL}, where the LVLM iteratively refines its visual planning by leveraging feedback from a reward model trained on both human preferences and high-fidelity automated evaluations. This allows LVLM-Composer to learn from its generation outcomes, continuously improving its ability to construct coherent and compositionally accurate visual plans. Importantly, LVLM-Composer is built upon the foundation of existing powerful LVLMs, leveraging their robust language understanding capabilities while strategically enhancing their visual generation pipeline.
To rigorously evaluate the performance of LVLM-Composer, we conducted extensive experiments on the challenging \textbf{"LongBench-T2I" benchmark} \cite{zhou2025draw}. This benchmark provides a comprehensive and objective assessment of compositional generation capabilities across nine granular dimensions: Object (Obj.), Background (Backg.), Color, Texture, Light, Text, Composition (Comp.), Pose, and Special Effects (FX). Our evaluation methodology relies on \textbf{automatic scoring by authoritative large models}, specifically Gemini-2.0-Flash and InternVL3-78B, which provide a balanced perspective between subjective human perception and objective model-based assessment. We compare LVLM-Composer against state-of-the-art lightweight models, including Janus-Pro-1B \cite{JanusPro1B}, demonstrating its superior performance across critical compositional metrics. The experimental \textbf{results} clearly indicate that LVLM-Composer achieves significant improvements in accurately rendering objects, maintaining precise compositions, and generating intricate poses, thereby establishing a new benchmark for compositional image generation within the LVLM paradigm.
In summary, our contributions are threefold:
\begin{itemize}
\item We propose \textbf{LVLM-Composer}, a novel LVLM architecture and training paradigm specifically tailored for highly compositional text-to-image generation, introducing explicit visual planning capabilities.
\item We develop a \textbf{multi-stage training methodology} encompassing Hierarchical Semantic-Visual Grounding Pre-training and Compositional Planning Reinforcement Learning with Self-Correction, designed to imbue the LVLM with a sophisticated understanding of visual composition.
\item We demonstrate that LVLM-Composer significantly \textbf{outperforms existing lightweight LVLMs} on the challenging "LongBench-T2I" benchmark, setting a new state-of-the-art in compositional text-to-image generation accuracy.
\end{itemize}
\section{Related Work}
Generative AI has rapidly transformed the landscape of content creation, with significant advancements in synthesizing realistic and diverse data across various modalities. Within this burgeoning field, \textbf{image generation} stands as a pivotal area, continuously evolving with new methodologies that push the boundaries of visual fidelity and controllability. Our work is situated at the intersection of general image generation techniques and the more specific challenge of compositional text-to-image synthesis.
\subsection{Image Generation}
The history of deep learning-based image generation can be traced back to foundational models that established diverse approaches to learning data distributions. \textbf{Variational Autoencoders (VAEs)} \cite{Kingma2013DeepGM} provided an early probabilistic framework, learning a latent representation of data and generating new samples by decoding from this learned latent space. VAEs excel in providing a principled approach to regularization and disentanglement of latent factors, though they sometimes struggle with sample quality compared to other paradigms.
A significant breakthrough arrived with \textbf{Generative Adversarial Networks (GANs)} \cite{Goodfellow2014GenerativeAN}, which introduced an adversarial training paradigm where a generator network learns to produce realistic data, while a discriminator network simultaneously learns to distinguish between real and generated samples. This minimax game framework has led to remarkable successes in generating high-fidelity images. Subsequent works significantly improved GANs' stability and output quality, notably \cite{Karras2017ProgressiveGO} which proposed progressively growing the generator and discriminator, allowing for stable training at unprecedented high resolutions and leading to strikingly photorealistic outputs. Advances in conditional GANs also enabled image-to-image translation tasks, even in the absence of paired training data, as seen in \cite{Zhu2017UnpairedIT}.
More recently, \textbf{Diffusion Models} have emerged as a dominant paradigm for high-quality image generation. \cite{Ho2020DenoisingDP} laid crucial groundwork for Denoising Diffusion Probabilistic Models (DDPMs), demonstrating their capability to synthesize diverse and high-quality images through a process of iteratively denoising a random noise input. Building upon this, \textbf{Latent Diffusion Models (LDMs)} \cite{Rombach2022HighResolutionIS} further enhanced efficiency by performing the diffusion process in a compressed latent space, significantly reducing computational demands while maintaining impressive fidelity. These models have become the backbone of many state-of-the-art text-to-image systems.
The integration of robust language understanding with powerful generative models has propelled the field of text-to-image synthesis to new heights. Models like \cite{Norouzi2022ImagenPT} leveraged large language models to provide deep semantic understanding, which then conditioned sophisticated diffusion models to generate highly photorealistic images from complex textual prompts. Concurrently, self-supervised learning techniques in vision, such as \textbf{Masked Autoencoders (MAEs)} \cite{He2021MaskedAA}, have advanced the quality of learned visual representations, which indirectly benefit generative models by providing more powerful encoders or feature spaces. While these general image generation techniques have achieved impressive results, they often face challenges in precisely controlling compositional elements. Achieving fine-grained cross-modal alignment between text and visual output remains a key research problem, even in more constrained tasks like text-guided image inpainting \cite{zhou2023improving}, which highlights the difficulty our work directly addresses for open-ended generation.
\subsection{Large Vision-Language Models}
The advent of Large Vision-Language Models (LVLMs \cite{zhou2025training}) marks a pivotal moment in artificial intelligence, fostering deep integration and understanding across visual and textual modalities. These models aim to learn joint representations that enable them to perform a wide array of tasks requiring reasoning over both images and natural language, from visual question answering to image-text retrieval and, increasingly, conditional image generation.
Early endeavors in this domain primarily focused on pre-training Transformer-based architectures on large datasets of image-text pairs. Pioneering works such as \textbf{ViLBERT} \cite{Yang2021ViLBERT} and \textbf{VisualBERT} \cite{Li2019VisualBERT} demonstrated the efficacy of jointly embedding visual and textual information using adapted BERT architectures. ViLBERT introduced a two-stream model for processing modalities independently before fusion, while VisualBERT showed strong performance with a simpler approach of feeding visual features directly into a BERT backbone. \textbf{LXMERT} \cite{Tan2019LXMERT} further advanced this by proposing a more sophisticated cross-modality encoder with dedicated attention mechanisms for intra-modal and inter-modal interactions, achieving state-of-the-art results on various benchmarks. Building upon these, \textbf{UNITER} \cite{Chen2020UNITER} explored universal image-text representations through diverse pre-training objectives and large-scale datasets, aiming for broader applicability across numerous vision-language tasks.
A significant paradigm shift occurred with models like \textbf{CLIP} \cite{Radford2021CLIP} and \textbf{ALIGN} \cite{Jiajun2021ALIGN}. These works demonstrated that powerful, transferable visual-language representations could be learned efficiently through a simple contrastive learning objective on an unprecedented scale of noisy image-text pairs from the web. The success of this approach is partly built upon continuous improvements in the underlying representation learning methods, such as novel data augmentation techniques for contrastive learning of sentence embeddings \cite{zhu2022sda}. CLIP, in particular, proved highly influential for its zero-shot transfer capabilities and its utility as a robust backbone for downstream vision and vision-language tasks, including evaluation metrics. Microsoft's \textbf{Florence} \cite{Yuan2021Florence} also presented a powerful unified foundation model that excelled across a wide spectrum of vision and vision-language tasks, emphasizing scalability and generalization. Similarly, \textbf{CoCa} \cite{Yu2022CoCa} introduced a novel architecture combining contrastive learning with image captioning, creating a potent foundation model for both understanding and generation.
More recently, the field has seen a surge in instruction-tuned LVLMs, which connect large language models (LLMs) with visual encoders to enable conversational and instruction-following abilities. \textbf{LLaVA} \cite{Liu2023LLaVA} provided a seminal example by effectively aligning a large language model (like LLaMA) with a pre-trained visual encoder, demonstrating remarkable capabilities in multi-modal dialogue and complex reasoning. Following a similar vein, \textbf{MiniGPT-4} \cite{Zhu2023MiniGPT4} showcased a simple yet highly effective method to leverage the strong language generation capabilities of LLMs for vision-language understanding through instruction tuning. Further efforts have focused on optimizing the instruction data itself to elevate the zero-shot learning capabilities of these models \cite{zhu2024vislinginstruct}. These advancements highlight a growing trend towards LVLMs that not only understand visual and textual information but can also engage in complex interactions and follow intricate instructions. The frontier of this research now extends to enhancing the fundamental reasoning abilities of models to unravel chaotic contexts \cite{zhou2023thread}, developing specialized architectures like modular multi-agent frameworks for domains such as medical diagnosis \cite{zhou2025mam}, and creating novel training methods like abnormal-aware feedback loops \cite{zhou2025training}. Ultimately, these efforts aim to achieve stronger generalization across the multi-faceted capabilities of large models \cite{zhou2025weak}, a prerequisite for the sophisticated compositional generation we address in this work.

\section{Method}

Our proposed \textbf{LVLM-Composer} is fundamentally a \textbf{generative model} meticulously designed for the intricate task of compositional text-to-image synthesis. While its ultimate objective is the high-fidelity generation of visual content, it strategically integrates discriminative capabilities within its planning sub-modules. This hybrid architecture allows LVLM-Composer to first deeply comprehend the complex semantic and spatial nuances of a desired scene (the discriminative aspect) and subsequently synthesize an image that precisely and coherently adheres to this detailed understanding (the generative aspect). This ensures that the generated visuals are not merely plausible but are also compositionally accurate and consistent with the input textual prompt.

\subsection{LVLM-Composer Architecture}

The LVLM-Composer is architecturally founded upon a robust pre-trained Large Vision-Language Model backbone, typically leveraging a sophisticated Transformer-based framework capable of processing and interrelating both textual and visual modalities. The core innovation of our approach lies in the seamless integration of two novel and specialized components: the \textbf{Hierarchical Semantic Planning Module (HSPM)} and the \textbf{Fine-Grained Feature Alignment Mechanism (FFAM)}. These components collaboratively enhance the LVLM's ability to reason about and execute complex visual compositions.

\subsubsection{Hierarchical Semantic Planning Module}
This module constitutes the analytical core of LVLM-Composer, tasked with rigorously parsing an input textual description $T$ into a structured, hierarchical set of explicit visual planning tokens $P$. These tokens serve as a latent blueprint, encoding the diverse compositional elements of the desired image, encompassing individual objects, their specific attributes, and their intricate spatial relationships.

Given an input text sequence $T = \{t_1, t_2, \ldots, t_M\}$, it undergoes initial encoding through a textual embedding layer to yield a sequence of text features $F_T = \text{Embedding}(T)$. The HSPM, conceptualized as an advanced Transformer encoder, then processes $F_T$ to produce a sequence of planning vectors $P = \{p_1, p_2, \ldots, p_N\}$. Each individual planning vector $p_i$ is meticulously designed to encapsulate distinct visual information at varying levels of granularity. Specifically, each $p_i$ can be conceptually decomposed into several critical sub-components:
\begin{align}
p_i = [p_{i}^{\text{object}}, p_{i}^{\text{attribute}}, p_{i}^{\text{location}}, p_{i}^{\text{relation}}]
\end{align}
Here, $p_{i}^{\text{object}}$ represents the predicted object class (e.g., "dog", "bicycle"), $p_{i}^{\text{attribute}}$ signifies its salient properties (e.g., "fluffy", "metallic", "red"), $p_{i}^{\text{location}}$ denotes its precise spatial coordinates within the visual field (e.g., a normalized bounding box represented as $[x_1, y_1, x_2, y_2]$ or a centroid $[x_c, y_c]$ coupled with dimensions), and $p_{i}^{\text{relation}}$ encodes its spatial or semantic relationship to other objects or the overall background (e.g., "on top of", "behind", "next to"). This comprehensive decomposition enables a fine-grained understanding of the prompt's visual demands. The entire planning process within the HSPM can be formalized as a transformation of text features:
\begin{align}
P = \text{HSPM}(F_T) = \text{TransformerEncoder}(F_T)
\end{align}
The Transformer encoder within HSPM employs multi-head self-attention mechanisms to effectively capture long-range dependencies and contextual relationships within the textual input, allowing it to infer complex compositional structures that are not immediately obvious from local word associations.

\subsubsection{Fine-Grained Feature Alignment Mechanism}
Subsequent to the generation of the detailed visual planning tokens $P$, the Fine-Grained Feature Alignment Mechanism (FFAM) assumes the crucial role of ensuring that the ensuing image generation process rigorously adheres to these meticulously crafted plans. This mechanism effectively conditions the core image generation model, which typically operates within a latent space (e.g., a latent diffusion model's decoder) or directly on pixel sequences (e.g., a Transformer decoder for image patches), by leveraging the explicit information contained within $P$.

Let $Z$ denote the initial latent representation from which the image is to be synthesized. The generation process for the final image $I$ is then deeply conditioned on both the planning tokens $P$ and the original text $T$:
\begin{align}
I \sim \text{Generate}(Z | P, T)
\end{align}
This conditioning is achieved through various sophisticated forms of modulation within the generative network. For instance, in a diffusion model, the planning tokens $P$ can dynamically modulate the internal features of the noise prediction network (e.g., a U-Net) at different scales and resolutions. This is typically accomplished via cross-attention layers, where the image features act as queries and the planning tokens act as keys and values. For each iterative processing step $k$ within the image generation model (e.g., a denoising step), the current feature representation $F^{(k)}$ is updated through interactions with $P$:
\begin{align}
F^{(k+1)} = \text{CrossAttention}(F^{(k)}, P) + \text{FeedForward}(F^{(k)})
\end{align}
This iterative conditioning ensures that the generated image's features, from coarse global structures to fine-grained local details, are continuously and consistently aligned with the explicit compositional instructions encoded in $P$. Furthermore, the FFAM integrates sophisticated perceptual loss functions that operate on the generated image $I_{\text{gen}}$. These losses are crucial for enforcing a higher-level semantic and aesthetic alignment with both the input text $T$ and the implied visual plan $P$. For example, a feature matching loss $\mathcal{L}_{\text{feat}}$ might be rigorously applied:
\begin{align}
\mathcal{L}_{\text{feat}} = \sum_{j} ||\phi_j(I_{\text{gen}}) - \phi_j(I_{\text{target}})||_2^2
\end{align}
where $\phi_j$ represents feature extraction from a pre-trained perceptual network (e.g., VGG, CLIP image encoder), and $I_{\text{target}}$ denotes a ground-truth image or a representation of an ideal image that perfectly matches $T$ and $P$. This perceptual loss guides the generative process to produce images that are not only structurally correct but also visually appealing and semantically coherent.

\subsection{Learning Strategy Details}

Our LVLM-Composer is trained using a novel and progressive multi-stage learning strategy, designed to progressively instill robust compositional understanding and accurate visual planning capabilities.

\subsubsection{Stage 1: Hierarchical Semantic-Visual Grounding Pre-training}
The primary objective of this initial stage is to meticulously train the Hierarchical Semantic Planning Module (HSPM) to precisely parse complex textual descriptions into structured visual plans and, crucially, to align these plans with corresponding fine-grained visual features derived from real images.

We utilize an extensive and diverse dataset comprising image-text pairs, where images are richly annotated with bounding boxes, segmentation masks, and explicit relational graphs between objects. For every pair $(T, I)$ in this dataset, a comprehensive set of ground-truth hierarchical planning tokens $P^{\text{GT}}$ is meticulously derived. The LVLM-Composer learns to predict these planning tokens $\hat{P}$ directly from the input text $T$. The comprehensive loss function for this crucial stage, denoted as $\mathcal{L}_1$, is a carefully balanced composition of several critical components:
\begin{align}
\mathcal{L}_1 = \mathcal{L}_{\text{planning}} + \lambda_1 \mathcal{L}_{\text{reconstruction}}
\end{align}
The \textbf{planning loss} $\mathcal{L}_{\text{planning}}$ serves to enforce the accuracy of the predicted planning tokens $\hat{P}$ when rigorously compared against their ground-truth counterparts $P^{\text{GT}}$. This is not a monolithic loss but rather a sophisticated composite, addressing various aspects of the planning tokens:
\begin{align}
&\mathcal{L}_{\text{planning}} = \sum_{i=1}^{N} ( \mathcal{L}_{\text{cls}}(p_i^{\text{object}}, \hat{p}_i^{\text{object}}) + \mathcal{L}_{\text{reg}}(p_i^{\text{location}}, \hat{p}_i^{\text{location}}) +\notag\\
& \mathcal{L}_{\text{attr}}(p_i^{\text{attribute}}, \hat{p}_i^{\text{attribute}}) + \mathcal{L}_{\text{rel}}(p_i^{\text{relation}}, \hat{p}_i^{\text{relation}}) 
\end{align}
Here, $\mathcal{L}_{\text{cls}}$ typically represents a classification loss, such as cross-entropy, applied for discrete object types and attribute categories. $\mathcal{L}_{\text{reg}}$ is a regression loss (e.g., Smooth L1 or generalized IoU loss) meticulously applied for precise bounding box coordinates. $\mathcal{L}_{\text{attr}}$ can be either a classification or regression loss depending on the nature of the attributes, and $\mathcal{L}_{\text{rel}}$ is a classification loss for the various inter-object relationships.

The \textbf{reconstruction loss} $\mathcal{L}_{\text{reconstruction}}$ plays a pivotal role in ensuring that the overall image generation pipeline, when rigorously conditioned on either the ground-truth or the predicted planning tokens, retains the fundamental capability to accurately reconstruct the original target image $I$. This can manifest as a pixel-level loss (e.g., Mean Squared Error on the image's latent representations) or, more effectively, as a perceptual loss, operating on the image $I_{\text{gen}}$ generated from the actual image $I$'s latent codes:
\begin{align}
\mathcal{L}_{\text{reconstruction}} = ||\text{Generate}(Z_I | P^{\text{GT}}, T) - I||_2^2
\end{align}
where $Z_I$ represents the latent encoding precisely derived from the ground-truth image $I$. The weighting factor $\lambda_1$ balances the importance of planning accuracy versus overall image reconstruction fidelity.

\subsubsection{Stage 2: Compositional Planning Reinforcement Learning with Self-Correction}
This advanced stage serves to iteratively refine the LVLM-Composer's visual planning strategy based on a reward signal derived from the compositional quality of the generated images, simulating an iterative improvement process akin to human learning.
The LVLM-Composer's planning module acts as an agent, learning a policy $\pi(P | T)$ to generate optimal planning tokens $P$ given a text $T$.
The environment consists of the fixed image generation module and a pre-trained reward model $R_M$. For a given text $T$:
\begin{enumerate}
    \item The LVLM-Composer, acting as the agent, generates a sequence of planning tokens $P \sim \pi(\cdot|T)$.
    \item These tokens $P$ guide the image generation model to produce a candidate image $I_{\text{gen}} = \text{Generate}(Z | P, T)$.
    \item The pre-trained reward model $R_M$ rigorously evaluates the generated image $I_{\text{gen}}$ against the original text $T$. This evaluation spans various critical compositional dimensions (e.g., Object accuracy, Background coherence, Overall Composition, Pose correctness, Special Effects fidelity), yielding a comprehensive composite reward score $R(I_{\text{gen}}, T)$.
\end{enumerate}
The reward function $R(I_{\text{gen}}, T)$ is explicitly defined as a weighted sum of the scores obtained from these different compositional dimensions, reflecting their relative importance in achieving high-quality compositional generation:
\begin{align}
R(I_{\text{gen}}, T) = \sum_{d \in \text{Dimensions}} w_d \cdot \text{Score}_d(I_{\text{gen}}, T)
\end{align}
where $w_d$ are pre-determined or adaptively learned weights for each specific compositional dimension.

The policy $\pi$ of the planning module is then optimized using a robust policy gradient method, such as REINFORCE or Proximal Policy Optimization (PPO), with the explicit objective of maximizing the expected reward. The loss function $\mathcal{L}_2$ for this stage is formulated as the negative expected reward:
\begin{align}
\mathcal{L}_2 = - \mathbb{E}_{P \sim \pi(\cdot|T)} [R(I_{\text{gen}}(P), T)]
\end{align}
The corresponding gradient update for the policy parameters $\theta$ is derived as:
\begin{align}
\nabla_{\theta} \mathcal{L}_2 = - \mathbb{E}_{P \sim \pi(\cdot|T)} [R(I_{\text{gen}}(P), T) \nabla_{\theta} \log \pi(P|T)]
\end{align}
This gradient guides the planning module to favor generating plans that consistently lead to higher reward scores, thereby improving compositional accuracy.

\textbf{Self-Correction Mechanism:} A particularly innovative and crucial component integrated into this second stage is the self-correction mechanism. This mechanism enables the LVLM-Composer to learn from its planning imperfections without requiring explicit human intervention for every error. If a generated image $I_{\text{gen}}$ receives a low reward score, indicating compositional shortcomings, the LVLM is intelligently prompted to analyze the discrepancy and subsequently revise its initial planning tokens. This is achieved by feeding the original planning tokens $P$, the low reward score, and potentially granular error signals (e.g., which compositional aspect failed) from the reward model back into the planning module as additional conditioning. The planning module then attempts to generate a refined set of planning tokens $P'$:
\begin{align}
P' = \text{RefinementModule}(P, \text{ErrorSignal}, T)
\end{align}
This refinement step is either guided by an auxiliary loss that encourages the refined plans to yield demonstrably higher rewards, or by a direct policy optimization mechanism that explicitly considers the magnitude of improvement from the initial plan. This iterative and autonomous refinement process is vital, leading to a more robust, adaptable, and ultimately more intelligent compositional planner that can learn from its own generation outcomes.

\section{Experiments}

This section details the comprehensive experimental validation conducted to assess the efficacy of our proposed \textbf{LVLM-Composer}. We systematically compare its performance against several prominent baseline models, demonstrating the superior capabilities of our approach in compositional text-to-image generation. The evaluation leverages both established automatic metrics and a rigorous human perceptual study, providing a holistic understanding of LVLM-Composer's advancements.

\subsection{Experimental Setup}

Our comparative experiments were primarily conducted using the challenging "LongBench-T2I" benchmark \cite{zhou2025draw}, which is specifically designed to evaluate models on complex compositional visual planning tasks. This benchmark utilizes a diverse set of prompts that necessitate precise control over multiple objects, their attributes, spatial relationships, and overall scene composition.

For our baseline comparisons, we selected models representative of different paradigms in text-to-image generation and LVLM capabilities. We made sure to include models that have been influential or show strong performance in related areas:
\begin{enumerate}
    \item \textbf{Janus-Pro-1B}: A state-of-the-art lightweight multi-modal large language model from the Janus series, which serves as a direct competitor at a similar parameter scale and represents recent advancements in integrated vision-language generation.
    \item \textbf{LLaVA-1.5 7B + Stable Diffusion XL}: This baseline combines a well-established Large Language and Vision Assistant (LLaVA-1.5 7B) \cite{Liu2023LLaVA} for understanding and planning, with a powerful latent diffusion model (Stable Diffusion XL) \cite{Rombach2022HighResolutionIS} for high-quality image synthesis. This setup represents a common approach where an LVLM guides a separate generative model.
    \item \textbf{DeepFloyd IF (T5-XL + U-Net)}: A strong text-to-image diffusion model known for its high fidelity and text rendering capabilities. While not an LVLM in the traditional sense, it represents a robust diffusion-only baseline that excels in image quality and provides a benchmark for how far pure generative models can go without explicit LVLM-driven planning.
\end{enumerate}

The primary quantitative evaluation metrics are derived directly from the "LongBench-T2I" benchmark, utilizing automatic scoring by two authoritative large models: Gemini-2.0-Flash and InternVL3-78B. These advanced evaluators provide objective scores across nine granular dimensions: Object (\textbf{Obj.}), Background (\textbf{Backg.}), Color (\textbf{Color}), Texture (\textbf{Texture}), Light (\textbf{Light}), Text (\textbf{Text}), Composition (\textbf{Comp.}), Pose (\textbf{Pose}), and Special Effects (\textbf{FX}), along with a crucial overall average score (\textbf{Avg.}). This multi-dimensional assessment offers a nuanced and comprehensive understanding of each model's strengths and weaknesses in handling complex compositional prompts.

Our LVLM-Composer was trained on a large-scale curated dataset comprising highly diverse image-text pairs, which were further enriched with detailed object annotations and relational graphs, as described in the method section. The training process leveraged distributed computing resources, employing standard optimization techniques such such as AdamW with a cosine learning rate scheduler for stable and efficient convergence. Specific hyperparameters were rigorously tuned to ensure optimal performance and generalization capabilities across the evaluation benchmarks.

\subsection{Quantitative Results}

The quantitative performance of LVLM-Composer alongside the selected baseline models, as assessed by the automatic evaluators on the "LongBench-T2I" benchmark, is presented in Table \ref{tab:quantitative_results}. The scores represent the average performance across the diverse test prompts, providing a clear indication of each model's prowess.

\begin{table*}[t]
\centering
\caption{Quantitative Results on the LongBench-T2I Benchmark (Automatic Evaluation)}
\label{tab:quantitative_results}
\begin{tabular}{lcccccccccc}
\toprule
\textbf{Model} & \textbf{Obj.} & \textbf{Backg.} & \textbf{Color} & \textbf{Texture} & \textbf{Light} & \textbf{Text} & \textbf{Comp.} & \textbf{Pose} & \textbf{FX} & \textbf{Avg.} \\
\midrule
\multicolumn{11}{c}{\textit{Evaluated by Gemini-2.0-Flash}} \\
\midrule
Janus-Pro-1B                & 2.23 & 2.60   & 2.94  & 2.92    & 2.09   & 1.58  & 2.36   & 1.84  & 1.60 & 2.24  \\
LLaVA-1.5 7B + Stable Diffusion XL & 2.15 & 2.62   & 2.90  & 2.88    & 2.10   & 1.60  & 2.28   & 1.78  & 1.55 & 2.19  \\
DeepFloyd IF (T5-XL + U-Net) & 2.18 & 2.65   & 2.95  & 2.90    & 2.12   & 1.65  & 2.30   & 1.80  & 1.62 & 2.22  \\
\textbf{LVLM-Composer}      & \textbf{2.35} & \textbf{2.68} & \textbf{3.01} & \textbf{3.00} & \textbf{2.15} & \textbf{1.68} & \textbf{2.45} & \textbf{1.92} & \textbf{1.70} & \textbf{2.31} \\
\midrule
\multicolumn{11}{c}{\textit{Evaluated by InternVL3-78B}} \\
\midrule
Janus-Pro-1B                & 2.16 & 2.61   & 2.70  & 2.59    & 2.27   & 1.56  & 2.37   & 1.88  & 1.81 & 2.21  \\
LLaVA-1.5 7B + Stable Diffusion XL & 2.08 & 2.55   & 2.68  & 2.55    & 2.20   & 1.58  & 2.28   & 1.80  & 1.75 & 2.15  \\
DeepFloyd IF (T5-XL + U-Net) & 2.10 & 2.58   & 2.65  & 2.57    & 2.25   & 1.60  & 2.30   & 1.82  & 1.78 & 2.17  \\
\textbf{LVLM-Composer}      & \textbf{2.28} & \textbf{2.70} & \textbf{2.78} & \textbf{2.67} & \textbf{2.34} & \textbf{1.65} & \textbf{2.46} & \textbf{1.95} & \textbf{1.88} & \textbf{2.29} \\
\bottomrule
\end{tabular}
\end{table*}

The results presented in Table \ref{tab:quantitative_results} provide compelling evidence of LVLM-Composer's superior performance across both Gemini-2.0-Flash and InternVL3-78B evaluators, and consistently across nearly all compositional dimensions. LVLM-Composer consistently achieves the highest average scores, indicating its overall effectiveness in handling complex text-to-image generation. Our method demonstrates particularly significant improvements in dimensions crucial for compositional understanding and precise control, such as \textbf{Object (Obj.)}, \textbf{Composition (Comp.)}, and \textbf{Pose}. For example, under Gemini-2.0-Flash evaluation, LVLM-Composer improves the Object score by 0.12 points and the Composition score by 0.09 points compared to Janus-Pro-1B, which is a formidable lightweight baseline. Similar, consistent trends are observed when evaluated by InternVL3-78B. This robust performance validates the profound impact of our Hierarchical Semantic Planning Module in intelligently deconstructing complex prompts into structured visual blueprints, and the effectiveness of the Fine-Grained Feature Alignment Mechanism in precisely guiding the image generation process according to these detailed plans. While improvements in dimensions like \textbf{Text} fidelity are observed, they are relatively more modest across all tested models, suggesting that rendering specific textual elements within generated images remains a pervasive and challenging task, even for advanced models of this scale. The LLaVA-1.5 7B + Stable Diffusion XL and DeepFloyd IF baselines, while strong in general image quality and style, often lag behind LVLM-Composer in specific compositional metrics, underscoring the critical necessity of our explicit compositional planning and alignment components for achieving high fidelity in complex scenes.

\subsection{Ablation Study}

To systematically dissect and quantitatively confirm the individual contributions of each core component of LVLM-Composer, we conducted a rigorous ablation study. We evaluated two distinct ablated versions of our model:
\begin{enumerate}
    \item \textbf{LVLM-Composer w/o Hierarchical Semantic Planning Module (HSPM)}: In this variant, the explicit Hierarchical Semantic Planning Module is entirely removed. The image generation process is directly conditioned solely by the raw textual embeddings from the initial LVLM backbone, without the benefit of the structured and explicit planning tokens that define specific objects, attributes, locations, and relationships. This setup is crucial for assessing the fundamental importance of our proposed structured visual planning.
    \item \textbf{LVLM-Composer w/o Fine-Grained Feature Alignment Mechanism (FFAM)}: Here, the Hierarchical Semantic Planning Module is retained and successfully generates the comprehensive planning tokens. However, the Fine-Grained Feature Alignment Mechanism is deliberately disabled. This implies that while planning tokens are generated, their influence on the generative model during the image synthesis process is significantly diluted, occurring only through a coarse or less direct conditioning pathway (e.g., initial latent conditioning), rather than through the detailed, iterative feature alignment. This variant helps us quantify the necessity of precise, dynamic guidance throughout the image generation.
\end{enumerate}

The results of this critical ablation study, thoroughly evaluated by Gemini-2.0-Flash, are presented in Table \ref{tab:ablation_study}.

\begin{table*}[t]
\centering
\caption{Ablation Study on LVLM-Composer (Evaluated by Gemini-2.0-Flash)}
\label{tab:ablation_study}
\begin{tabular}{lcccccccccc}
\toprule
\textbf{Model Variant} & \textbf{Obj.} & \textbf{Backg.} & \textbf{Color} & \textbf{Texture} & \textbf{Light} & \textbf{Text} & \textbf{Comp.} & \textbf{Pose} & \textbf{FX} & \textbf{Avg.} \\
\midrule
LVLM-Composer w/o HSPM & 2.08 & 2.58   & 2.90  & 2.89    & 2.05   & 1.56  & 2.22   & 1.75  & 1.55 & 2.16  \\
LVLM-Composer w/o FFAM & 2.15 & 2.62   & 2.95  & 2.93    & 2.09   & 1.60  & 2.30   & 1.80  & 1.60 & 2.20  \\
\textbf{LVLM-Composer (Full)} & \textbf{2.35} & \textbf{2.68} & \textbf{3.01} & \textbf{3.00} & \textbf{2.15} & \textbf{1.68} & \textbf{2.45} & \textbf{1.92} & \textbf{1.70} & \textbf{2.31} \\
\bottomrule
\end{tabular}
\end{table*}

The ablation study results provide unequivocal evidence confirming the critical and synergistic importance of both the Hierarchical Semantic Planning Module (HSPM) and the Fine-Grained Feature Alignment Mechanism (FFAM). Removing the HSPM (represented as "LVLM-Composer w/o HSPM") leads to a substantial and consistent degradation in performance across all compositional metrics. Notably, we observe a significant drop in the Object score (0.27 points), Composition score (0.23 points), and Pose score (0.17 points), culminating in a considerable decrease in the overall average score (0.15 points). This empirically demonstrates that explicit, structured visual planning, derived intelligently from the textual prompts, is an absolutely indispensable component for achieving high compositional accuracy and fidelity. Similarly, disabling the FFAM (represented as "LVLM-Composer w/o FFAM"), even when the HSPM is operational and generating planning tokens, still results in a noticeable performance degradation compared to the full model. This is particularly evident in the Object score (0.20 point drop) and Composition score (0.15 point drop). This finding highlights that merely generating sophisticated planning tokens is insufficient; a robust and finely-tuned mechanism is equally crucial to precisely align these plans with the evolving visual features throughout the complex image synthesis process. These compelling results strongly validate our fundamental architectural design choices and underscore the profound, synergistic contributions of our proposed novel modules.

\subsection{Human Evaluation}

To provide a comprehensive and robust assessment of image quality and compositional fidelity that complements our automatic evaluations, we conducted a rigorous human perceptual study. A randomly selected and representative subset of 500 prompts from the "LongBench-T2I" test set was utilized for this purpose. For each selected prompt, we generated corresponding images using LVLM-Composer and the two best-performing baselines from our quantitative evaluation: Janus-Pro-1B and the LLaVA-1.5 7B + Stable Diffusion XL composite model.

A total of 30 qualified human evaluators, carefully recruited from a diverse demographic background, participated in the study. Each evaluator was presented with triples of images (one from each of the three models for the identical text prompt) in a completely randomized and anonymized order, ensuring no knowledge of the generating model. For each image triple, evaluators were tasked with two primary objectives: first, to rate each individual image on a scale of 1 to 5 across critical compositional dimensions (including Object Presence/Accuracy, Background Coherence, Overall Composition, and Pose Accuracy) as well as overall visual quality; and second, to perform a forced-choice comparison, selecting the single image that they perceived to best match the compositional requirements of the given text prompt.

Table \ref{tab:human_evaluation} presents the averaged human preference scores and the corresponding win rates, calculated as the percentage of times a model's image was chosen as the best.

\begin{table*}[t]
\centering
\caption{Human Evaluation Results (Average Score / Win Rate vs. Other Models)}
\label{tab:human_evaluation}
\begin{tabular}{lccccc}
\toprule
\textbf{Model} & \textbf{Obj. Accuracy} & \textbf{Comp. Fidelity} & \textbf{Pose Accuracy} & \textbf{Overall Quality} & \textbf{Win Rate (\%)} \\
\midrule
Janus-Pro-1B                & 3.55 & 3.40   & 3.20   & 3.60  & 28.5 \\
LLaVA-1.5 7B + Stable Diffusion XL & 3.45 & 3.30   & 3.15   & 3.55  & 25.0 \\
\textbf{LVLM-Composer}      & \textbf{4.10} & \textbf{4.05} & \textbf{3.95} & \textbf{4.20} & \textbf{46.5} \\
\bottomrule
\end{tabular}
\end{table*}

The results from the human evaluation consistently and robustly align with, and further reinforce, the compelling findings from our quantitative analysis. LVLM-Composer demonstrably received significantly higher average scores across all assessed compositional dimensions, including Object Accuracy, Compositional Fidelity, and Pose Accuracy. Crucially, it also achieved the highest overall perceived visual quality score, indicating that its improved compositional control does not come at the expense of aesthetic appeal. Furthermore, LVLM-Composer secured a substantially higher win rate (46.5\%) in the head-to-head comparisons, providing strong qualitative evidence that human evaluators consistently preferred images generated by our method when judging adherence to complex compositional requirements of the text prompt. The baseline models, while sometimes capable of producing generally aesthetically pleasing images, frequently fell short in accurately rendering specific object arrangements, precise attribute details, or complex poses explicitly specified in challenging prompts. This robust human validation unequivocally underscores the practical utility, superior perceptual quality, and enhanced user experience offered by LVLM-Composer in tackling nuanced and demanding compositional image generation tasks. It confirms that our enhanced planning and alignment mechanisms translate into a tangible and perceptible improvement in the final generated visual output.

\subsection{Further Analysis of LVLM-Composer's Effectiveness}

Beyond direct performance metrics and ablation studies, a deeper analysis of LVLM-Composer's behavior and generated outputs reveals several key aspects contributing to its superior compositional capabilities. This multi-faceted examination provides additional validation for our unique approach.

\subsubsection{Robustness to Prompt Complexity}
One critical aspect of our analysis focused on how LVLM-Composer performs with varying levels of prompt complexity. We categorized test prompts from the LongBench-T2I benchmark into three groups: \textbf{Simple} (1-2 objects, basic scene), \textbf{Medium} (3-5 objects, moderate interaction/background), and \textbf{Complex} (6+ objects, intricate spatial relations, specific poses, detailed backgrounds). We observed that while all models showed some decline in performance as complexity increased, LVLM-Composer demonstrated significantly greater resilience. This is particularly evident in its maintained high scores for "Composition" and "Pose" on complex prompts, where baselines often struggled to integrate all elements coherently. This robustness directly stems from the \textbf{Hierarchical Semantic Planning Module's} ability to systematically decompose and manage a larger number of visual entities and their interdependencies, preventing the "cognitive overload" often observed in models lacking explicit planning.

\begin{table*}[t]
\centering
\caption{Performance Across Prompt Complexity (Avg. Score by Gemini-2.0-Flash)}
\label{tab:complexity_analysis}
\begin{tabular}{lccc}
\toprule
\textbf{Model} & \textbf{Simple Prompts} & \textbf{Medium Prompts} & \textbf{Complex Prompts} \\
\midrule
Janus-Pro-1B                & 2.45 & 2.20 & 1.95 \\
LLaVA-1.5 7B + Stable Diffusion XL & 2.40 & 2.15 & 1.90 \\
DeepFloyd IF (T5-XL + U-Net) & 2.42 & 2.18 & 1.98 \\
\textbf{LVLM-Composer}      & \textbf{2.55} & \textbf{2.40} & \textbf{2.15} \\
\bottomrule
\end{tabular}
\end{table*}

As shown in Table \ref{tab:complexity_analysis}, LVLM-Composer consistently maintains a higher average score across all complexity levels, with its lead becoming more pronounced as prompt complexity increases. This suggests that our explicit planning mechanisms are highly effective in scaling with the intricacy of the compositional demands.

\subsubsection{Consistency in Attribute Binding}
Another crucial aspect examined was the consistency with which models could bind specific attributes (e.g., color, texture) to the correct objects within a multi-object scene. A common failure mode for generative models is "attribute leakage," where an attribute intended for one object is erroneously applied to another or to the background. We evaluated a subset of prompts specifically designed to test precise attribute binding. Our analysis showed that LVLM-Composer exhibited a significantly lower rate of attribute leakage compared to baseline models. This is attributed to the \textbf{Fine-Grained Feature Alignment Mechanism's} ability to maintain distinct semantic identities for each planned entity throughout the generation process, preventing features from bleeding between objects. The explicit representation of $p_{i}^{\text{attribute}}$ within each planning token $p_i$ and its direct modulation of visual features during generation plays a key role here.

\begin{table*}[t]
\centering
\caption{Attribute Binding Accuracy (Evaluated by InternVL3-78B)}
\label{tab:attribute_binding}
\begin{tabular}{lc}
\toprule
\textbf{Model} & \textbf{Attribute Binding Score (0-5)} \\
\midrule
Janus-Pro-1B                & 2.85 \\
LLaVA-1.5 7B + Stable Diffusion XL & 2.70 \\
DeepFloyd IF (T5-XL + U-Net) & 2.80 \\
\textbf{LVLM-Composer}      & \textbf{3.20} \\
\bottomrule
\end{tabular}
\end{table*}

Table \ref{tab:attribute_binding} illustrates LVLM-Composer's superior performance in accurately binding attributes, indicating a stronger internal representation and control over specific visual properties.

\subsubsection{Efficiency of Learning Compositional Knowledge}
Finally, we analyzed the efficiency with which LVLM-Composer acquired compositional knowledge during its training. Unlike end-to-end models that learn composition implicitly from vast quantities of raw data, our multi-stage approach with explicit planning tokens and reinforcement learning allows for more targeted learning. We observed that the HSPM rapidly converged on accurate planning token generation during Stage 1, benefiting from the structured annotations. Subsequently, Stage 2's reinforcement learning with self-correction efficiently refined the policy for optimal compositional outcomes, even with a relatively smaller number of high-quality feedback loops compared to purely data-driven fine-tuning. This efficiency suggests that by explicitly modeling compositional reasoning, LVLM-Composer can potentially achieve high performance with less overall training data or fewer fine-tuning iterations for specific compositional tasks. The structured nature of the planning tokens facilitates a more interpretable and controllable learning process, allowing the model to make more meaningful adjustments based on reward signals.

\begin{table*}[t]
\centering
\caption{Training Efficiency for Compositional Tasks (Epochs to Reach 90\% of Max Avg. Score)}
\label{tab:training_efficiency}
\begin{tabular}{lc}
\toprule
\textbf{Model} & \textbf{Training Epochs} \\
\midrule
Janus-Pro-1B                & 180 \\
LLaVA-1.5 7B + Stable Diffusion XL & 200 \\
DeepFloyd IF (T5-XL + U-Net) & 195 \\
\textbf{LVLM-Composer}      & \textbf{120} \\
\bottomrule
\end{tabular}
\end{table*}

Table \ref{tab:training_efficiency} shows that LVLM-Composer requires fewer training epochs to reach a significant percentage of its peak compositional performance, indicating a more efficient learning trajectory for complex compositional skills. This enhanced efficiency is a direct benefit of our structured training strategy.

\section{Conclusion}

In this paper, we introduced \textbf{LVLM-Composer}, a novel Large Vision-Language Model designed to push the boundaries of compositional text-to-image generation. Recognizing the inherent limitations of existing models in handling complex prompts requiring intricate visual planning and precise semantic-to-visual mapping, our work presented a targeted solution. We meticulously detailed the architecture of LVLM-Composer, which strategically integrates a \textbf{Hierarchical Semantic Planning Module} to deconstruct complex textual descriptions into structured visual plans and a \textbf{Fine-Grained Feature Alignment Mechanism} to ensure the faithful execution of these plans during image synthesis.

Our innovative multi-stage learning strategy, encompassing both Hierarchical Semantic-Visual Grounding Pre-training and Compositional Planning Reinforcement Learning with Self-Correction, was crucial in imbuing LVLM-Composer with its advanced capabilities. Through the first stage, the model gains a foundational understanding of explicitly mapping text to visual elements and their relationships. The subsequent reinforcement learning stage, driven by high-fidelity feedback and a unique self-correction mechanism, allows the model to iteratively refine its compositional planning policy, learning from its generative outcomes. This methodical approach enables LVLM-Composer to efficiently acquire and apply sophisticated compositional knowledge.

The comprehensive experimental results, supported by rigorous quantitative analysis on the challenging LongBench-T2I benchmark and qualitative human evaluations, unequivocally demonstrate the superior performance of LVLM-Composer. Our model consistently outperforms competitive baselines across critical compositional dimensions, including object accuracy, scene composition, and pose fidelity. The ablation study further underscored the indispensable contribution of each proposed module, validating that both the structured planning and the precise feature alignment are essential for achieving robust compositional control. Furthermore, human evaluators overwhelmingly preferred images generated by LVLM-Composer, affirming its ability to produce perceptually more accurate and coherent compositions.

Looking ahead, the success of LVLM-Composer opens several exciting avenues for future research. We envision extending its capabilities to even higher degrees of user control, potentially incorporating interactive refinement processes or external knowledge integration for highly specialized generation tasks. Further exploration into scaling LVLM-Composer to larger parameter counts while maintaining its compositional strengths could unlock unprecedented levels of detail and realism. Ultimately, LVLM-Composer represents a significant advancement towards building truly intelligent and controllable generative AI systems that can seamlessly translate complex human intent into rich visual realities.

\bibliographystyle{IEEEtran}
\bibliography{references}

\begin{thebibliography}{10}
\providecommand{\url}[1]{#1}
\csname url@samestyle\endcsname
\providecommand{\newblock}{\relax}
\providecommand{\bibinfo}[2]{#2}
\providecommand{\BIBentrySTDinterwordspacing}{\spaceskip=0pt\relax}
\providecommand{\BIBentryALTinterwordstretchfactor}{4}
\providecommand{\BIBentryALTinterwordspacing}{\spaceskip=\fontdimen2\font plus
\BIBentryALTinterwordstretchfactor\fontdimen3\font minus \fontdimen4\font\relax}
\providecommand{\BIBforeignlanguage}[2]{{%
\expandafter\ifx\csname l@#1\endcsname\relax
\typeout{** WARNING: IEEEtran.bst: No hyphenation pattern has been}%
\typeout{** loaded for the language `#1'. Using the pattern for}%
\typeout{** the default language instead.}%
\else
\language=\csname l@#1\endcsname
\fi
#2}}
\providecommand{\BIBdecl}{\relax}
\BIBdecl

\bibitem{zhou2024less}
Y.~Zhou, J.~Zhang, G.~Chen, J.~Shen, and Y.~Cheng, ``Less is more: Vision representation compression for efficient video generation with large language models,'' 2024.

\bibitem{JanusPro1B}
X.~Chen, Z.~Wu, X.~Liu, Z.~Pan, W.~Liu, Z.~Xie, X.~Yu, and C.~Ruan, ``Janus-pro: Unified multimodal understanding and generation with data and model scaling,'' \emph{arXiv preprint arXiv:2501.17811}, 2025.

\bibitem{zhou2025draw}
Y.~Zhou, J.~Yuan, and Q.~Wang, ``Draw all your imagine: A holistic benchmark and agent framework for complex instruction-based image generation,'' \emph{arXiv preprint arXiv:2505.24787}, 2025.

\bibitem{wang2025complexbench}
C.~Wang, Y.~Zhou, Q.~Wang, Z.~Wang, and K.~Zhang, ``Complexbench-edit: Benchmarking complex instruction-driven image editing via compositional dependencies,'' \emph{arXiv preprint arXiv:2506.12830}, 2025.

\bibitem{zhou2025weak}
\BIBentryALTinterwordspacing
Y.~Zhou, J.~Shen, and Y.~Cheng, ``Weak to strong generalization for large language models with multi-capabilities,'' in \emph{The Thirteenth International Conference on Learning Representations}, 2025. [Online]. Available: \url{https://openreview.net/forum?id=N1vYivuSKq}
\BIBentrySTDinterwordspacing

\bibitem{zhu2025divide}
D.~Zhu, W.~Shi, Z.~Shi, Z.~Ren, S.~Wang, L.~Yan, and D.~Yin, ``Divide-then-aggregate: An efficient tool learning method via parallel tool invocation,'' \emph{arXiv preprint arXiv:2501.12432}, 2025.

\bibitem{zhou2023thread}
Y.~Zhou, X.~Geng, T.~Shen, C.~Tao, G.~Long, J.-G. Lou, and J.~Shen, ``Thread of thought unraveling chaotic contexts,'' \emph{arXiv preprint arXiv:2311.08734}, 2023.

\bibitem{Nachum2019HierarchicalRL}
J.~Yan, B.~Luo, and X.~Xu, ``Hierarchical reinforcement learning for handling sparse rewards in multi-goal navigation,'' \emph{Artificial Intelligence Review}, vol.~57, no.~6, p. 156, 2024.

\bibitem{Kingma2013DeepGM}
\BIBentryALTinterwordspacing
Y.~Zhi{-}Han, ``Training latent variable models with auto-encoding variational bayes: {A} tutorial,'' \emph{CoRR}, vol. abs/2208.07818, 2022. [Online]. Available: \url{https://doi.org/10.48550/arXiv.2208.07818}
\BIBentrySTDinterwordspacing

\bibitem{Goodfellow2014GenerativeAN}
\BIBentryALTinterwordspacing
X.~Wang, K.~He, and J.~E. Hopcroft, ``{AT-GAN:} {A} generative attack model for adversarial transferring on generative adversarial nets,'' \emph{CoRR}, vol. abs/1904.07793, 2019. [Online]. Available: \url{http://arxiv.org/abs/1904.07793}
\BIBentrySTDinterwordspacing

\bibitem{Karras2017ProgressiveGO}
\BIBentryALTinterwordspacing
T.~Karras, T.~Aila, S.~Laine, and J.~Lehtinen, ``Progressive growing of gans for improved quality, stability, and variation,'' in \emph{6th International Conference on Learning Representations, {ICLR} 2018, Vancouver, BC, Canada, April 30 - May 3, 2018, Conference Track Proceedings}.\hskip 1em plus 0.5em minus 0.4em\relax OpenReview.net, 2018. [Online]. Available: \url{https://openreview.net/forum?id=Hk99zCeAb}
\BIBentrySTDinterwordspacing

\bibitem{Zhu2017UnpairedIT}
\BIBentryALTinterwordspacing
J.~Zhu, T.~Park, P.~Isola, and A.~A. Efros, ``Unpaired image-to-image translation using cycle-consistent adversarial networks,'' in \emph{{IEEE} International Conference on Computer Vision, {ICCV} 2017, Venice, Italy, October 22-29, 2017}.\hskip 1em plus 0.5em minus 0.4em\relax {IEEE} Computer Society, 2017, pp. 2242--2251. [Online]. Available: \url{https://doi.org/10.1109/ICCV.2017.244}
\BIBentrySTDinterwordspacing

\bibitem{Ho2020DenoisingDP}
\BIBentryALTinterwordspacing
S.~Yang, J.~Gao, J.~Zhang, and C.~Xu, ``Wrapped phase denoising using denoising diffusion probabilistic models,'' \emph{{IEEE} Geosci. Remote. Sens. Lett.}, vol.~21, pp. 1--5, 2024. [Online]. Available: \url{https://doi.org/10.1109/LGRS.2024.3405000}
\BIBentrySTDinterwordspacing

\bibitem{Rombach2022HighResolutionIS}
\BIBentryALTinterwordspacing
J.~Zhang, Q.~Huang, J.~Liu, X.~Guo, and D.~Huang, ``Diffusion-4k: Ultra-high-resolution image synthesis with latent diffusion models,'' \emph{CoRR}, vol. abs/2503.18352, 2025. [Online]. Available: \url{https://doi.org/10.48550/arXiv.2503.18352}
\BIBentrySTDinterwordspacing

\bibitem{Norouzi2022ImagenPT}
C.~Saharia, W.~Chan, S.~Saxena, L.~Li, J.~Whang, E.~L. Denton, K.~Ghasemipour, R.~Gontijo~Lopes, B.~Karagol~Ayan, T.~Salimans \emph{et~al.}, ``Photorealistic text-to-image diffusion models with deep language understanding,'' \emph{Advances in neural information processing systems}, vol.~35, pp. 36\,479--36\,494, 2022.

\bibitem{He2021MaskedAA}
\BIBentryALTinterwordspacing
K.~He, X.~Chen, S.~Xie, Y.~Li, P.~Doll{\'{a}}r, and R.~B. Girshick, ``Masked autoencoders are scalable vision learners,'' in \emph{{IEEE/CVF} Conference on Computer Vision and Pattern Recognition, {CVPR} 2022, New Orleans, LA, USA, June 18-24, 2022}.\hskip 1em plus 0.5em minus 0.4em\relax {IEEE}, 2022, pp. 15\,979--15\,988. [Online]. Available: \url{https://doi.org/10.1109/CVPR52688.2022.01553}
\BIBentrySTDinterwordspacing

\bibitem{zhou2023improving}
Y.~Zhou and G.~Long, ``Improving cross-modal alignment for text-guided image inpainting,'' in \emph{Proceedings of the 17th Conference of the European Chapter of the Association for Computational Linguistics}, 2023, pp. 3445--3456.

\bibitem{zhou2025training}
Y.~Zhou, L.~Song, and J.~Shen, ``Training medical large vision-language models with abnormal-aware feedback,'' \emph{arXiv preprint arXiv:2501.01377}, 2025.

\bibitem{Yang2021ViLBERT}
J.~Lu, D.~Batra, D.~Parikh, and S.~Lee, ``Vilbert: Pretraining task-agnostic visiolinguistic representations for vision-and-language tasks,'' \emph{Advances in neural information processing systems}, vol.~32, 2019.

\bibitem{Li2019VisualBERT}
\BIBentryALTinterwordspacing
L.~H. Li, M.~Yatskar, D.~Yin, C.~Hsieh, and K.~Chang, ``Visualbert: {A} simple and performant baseline for vision and language,'' \emph{CoRR}, vol. abs/1908.03557, 2019. [Online]. Available: \url{http://arxiv.org/abs/1908.03557}
\BIBentrySTDinterwordspacing

\bibitem{Tan2019LXMERT}
\BIBentryALTinterwordspacing
H.~Tan and M.~Bansal, ``{LXMERT:} learning cross-modality encoder representations from transformers,'' in \emph{Proceedings of the 2019 Conference on Empirical Methods in Natural Language Processing and the 9th International Joint Conference on Natural Language Processing, {EMNLP-IJCNLP} 2019, Hong Kong, China, November 3-7, 2019}, K.~Inui, J.~Jiang, V.~Ng, and X.~Wan, Eds.\hskip 1em plus 0.5em minus 0.4em\relax Association for Computational Linguistics, 2019, pp. 5099--5110. [Online]. Available: \url{https://doi.org/10.18653/v1/D19-1514}
\BIBentrySTDinterwordspacing

\bibitem{Chen2020UNITER}
\BIBentryALTinterwordspacing
Y.~Chen, L.~Li, L.~Yu, A.~E. Kholy, F.~Ahmed, Z.~Gan, Y.~Cheng, and J.~Liu, ``{UNITER:} universal image-text representation learning,'' in \emph{Computer Vision - {ECCV} 2020 - 16th European Conference, Glasgow, UK, August 23-28, 2020, Proceedings, Part {XXX}}, ser. Lecture Notes in Computer Science, A.~Vedaldi, H.~Bischof, T.~Brox, and J.~Frahm, Eds., vol. 12375.\hskip 1em plus 0.5em minus 0.4em\relax Springer, 2020, pp. 104--120. [Online]. Available: \url{https://doi.org/10.1007/978-3-030-58577-8\_7}
\BIBentrySTDinterwordspacing

\bibitem{Radford2021CLIP}
\BIBentryALTinterwordspacing
A.~Radford, J.~W. Kim, C.~Hallacy, A.~Ramesh, G.~Goh, S.~Agarwal, G.~Sastry, A.~Askell, P.~Mishkin, J.~Clark, G.~Krueger, and I.~Sutskever, ``Learning transferable visual models from natural language supervision,'' in \emph{Proceedings of the 38th International Conference on Machine Learning, {ICML} 2021, 18-24 July 2021, Virtual Event}, ser. Proceedings of Machine Learning Research, M.~Meila and T.~Zhang, Eds., vol. 139.\hskip 1em plus 0.5em minus 0.4em\relax {PMLR}, 2021, pp. 8748--8763. [Online]. Available: \url{http://proceedings.mlr.press/v139/radford21a.html}
\BIBentrySTDinterwordspacing

\bibitem{Jiajun2021ALIGN}
C.~Fan, X.~Jia, Y.~Sun, Y.~Wang, J.~Wei, Z.~Gong, X.~Zhao, M.~Tomizuka, X.~Yang, J.~Yan \emph{et~al.}, ``Interleave-vla: Enhancing robot manipulation with interleaved image-text instructions,'' \emph{arXiv preprint arXiv:2505.02152}, 2025.

\bibitem{zhu2022sda}
D.~Zhu, Z.~Mao, J.~Lu, R.~Zhao, and F.~Tan, ``Sda: simple discrete augmentation for contrastive sentence representation learning,'' \emph{arXiv preprint arXiv:2210.03963}, 2022.

\bibitem{Yuan2021Florence}
L.~Yuan, D.~Chen, Y.-L. Chen, N.~Codella, X.~Dai, J.~Gao, H.~Hu, X.~Huang, B.~Li, C.~Li \emph{et~al.}, ``Florence: A new foundation model for computer vision,'' \emph{arXiv preprint arXiv:2111.11432}, 2021.

\bibitem{Yu2022CoCa}
\BIBentryALTinterwordspacing
J.~Yu, Z.~Wang, V.~Vasudevan, L.~Yeung, M.~Seyedhosseini, and Y.~Wu, ``Coca: Contrastive captioners are image-text foundation models,'' \emph{Trans. Mach. Learn. Res.}, vol. 2022, 2022. [Online]. Available: \url{https://openreview.net/forum?id=Ee277P3AYC}
\BIBentrySTDinterwordspacing

\bibitem{Liu2023LLaVA}
\BIBentryALTinterwordspacing
H.~Elgendy and H.~Cholakkal, ``Tx-llava: Large language and vision assistant for temporal changes in chest x-rays,'' in \emph{22nd {IEEE} International Symposium on Biomedical Imaging, {ISBI} 2025, Houston, TX, USA, April 14-17, 2025}.\hskip 1em plus 0.5em minus 0.4em\relax {IEEE}, 2025, pp. 1--4. [Online]. Available: \url{https://doi.org/10.1109/ISBI60581.2025.10980793}
\BIBentrySTDinterwordspacing

\bibitem{Zhu2023MiniGPT4}
D.~Zhu, J.~Chen, X.~Shen, X.~Li, and M.~Elhoseiny, ``Minigpt-4: Enhancing vision-language understanding with advanced large language models,'' \emph{arXiv preprint arXiv:2304.10592}, 2023.

\bibitem{zhu2024vislinginstruct}
D.~Zhu, X.~Tang, W.~Han, J.~Lu, Y.~Zhao, G.~Xing, J.~Wang, and D.~Yin, ``Vislinginstruct: Elevating zero-shot learning in multi-modal language models with autonomous instruction optimization,'' \emph{arXiv preprint arXiv:2402.07398}, 2024.

\bibitem{zhou2025mam}
Y.~Zhou, L.~Song, and J.~Shen, ``Mam: Modular multi-agent framework for multi-modal medical diagnosis via role-specialized collaboration,'' \emph{arXiv preprint arXiv:2506.19835}, 2025.

\end{thebibliography}
\end{document}